\documentclass[conference]{IEEEtran}
\IEEEoverridecommandlockouts
\usepackage{cite}
\usepackage{array}
\usepackage{makecell}
\usepackage{multirow}
\usepackage{amsmath,amssymb,amsfonts}
\usepackage{algorithmic}
\usepackage{graphicx}
\usepackage{epstopdf}
\usepackage{textcomp}
\usepackage{xcolor}
\def\BibTeX{{\rm B\kern-.05em{\sc i\kern-.025em b}\kern-.08em
    T\kern-.1667em\lower.7ex\hbox{E}\kern-.125emX}}

\begin{document}
\renewcommand\arraystretch{1.5}		% 将表格高度设置为默认值的x倍

% 题目 资助项目
\title{SGDN: Segmentation-Based Grasp Detection Network For Unsymmetrical Three-Finger Gripper \\
\thanks{This work was supported by the National Key Research and Development Project of China, No. 2018YFB1305300, and Shandong Provincial Natural Science Foundation of China, No. 2019JZZY010130 and 2018CXGC0907.}
}

% 作者
\author{
	\IEEEauthorblockN{WANG Dexin}
	\IEEEauthorblockA{
		\textit{School of Control Science and Engineering} \\
		\textit{Shandong University}\\
		Jinan, China \\
		dexinwang@mail.sdu.edu.cn}
	\and
	\IEEEauthorblockN{CHANG Faliang}
	\IEEEauthorblockA{
		\textit{School of Control Science and Engineering} \\
		\textit{Shandong University}\\
		Jinan, China}
	\and
	\IEEEauthorblockN{LIU Chunsheng}
	\IEEEauthorblockA{
		\textit{School of Control Science and Engineering} \\
		\textit{Shandong University}\\
		Jinan, China}
	\and
	\IEEEauthorblockN{LI Nanjun}
	\IEEEauthorblockA{
		\textit{School of Control Science and Engineering} \\
		\textit{Shandong University}\\
		Jinan, China}
	
}

% 显示导言信息
\maketitle

% 摘要
\begin{abstract}
In this paper, we present Segmentation-Based Grasp Detection Network (SGDN) to predict a feasible robotic grasping for a unsymmetrical three-finger robotic gripper using RGB images. 
The feasible grasping of a target should be a collection of grasp regions with the same grasp angle and width. 
In other words, a simplified planar grasp representation should be pixel-level rather than region-level such as five-dimensional grasp representation.
Therefore, we propose a pixel-level grasp representation, oriented base-fixed triangle. 
It is also more suitable for unsymmetrical three-finger gripper which cannot grasp symmetrically when grasping some objects, the grasp angle is at $[0, 2\pi)$ instead of $[0, \pi)$ of parallel plate gripper.
In order to predict the appropriate grasp region and its corresponding grasp angle and width in the RGB image, SGDN uses DeepLabv3+ as a feature extractor, and uses a three-channel grasp predictor to predict feasible oriented base-fixed triangle grasp representation of each pixel.
On the re-annotated Cornell Grasp Dataset, our model achieves an accuracy of 96.8\% and 92.27\% on image-wise split and object-wise split respectively, and obtains accurate predictions consistent with the state-of-the-art methods.
\end{abstract}

% 关键词 

\section{Introduction}
% 图1 流程图
\begin{figure}[htbp]
	\centerline
	{\includegraphics[scale=0.6]{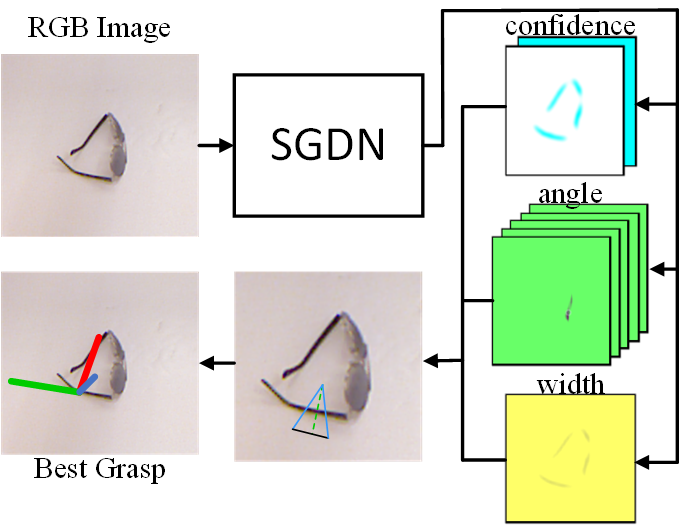}}
	\caption{Our real-time, generative grasping pipeline. After the camera installed in the scene captures the RGB image containing the object to be grasped, our Segment-Based Grasp Detection Network (SGDN) generates grasping – parameterised as a grasp position, angle and gripper width – for every pixel in the input RGB image in a fraction of a second. The best grasping is calculated and robot performs grasping.}
	\label{fig1}
\end{figure}
In household and industrial scenes, grasp objects from the table is a very important ability when the robot is running independently or when human-machine cooperation.
Even children without education can instinctively grasp objects accurately, stably and quickly, but for robots, grasping is still a huge challenge.
The robot grasp problems can be divided into grasp detection, trajectory planning, and execution.
Among them, grasp detection is both the first step and the most important step. The robot obtains the visual information of the target to be grasped through the RGB or RGBD camera, and then uses these visual information to detect the grasp model that can be used to implement the grasping.
The grasp model is mapped to the coordinates in the real world through the transformation matrix between the pixel coordinate system and the world coordinate system, and the robot can open the gripper, plan the trajectory and move the gripper to the object, and then the gripper closes to grasp.
In many scenes, we fix the robot next to the table, and the camera is fixed in the scene or installed above the wrist of the robotic arm. 
If the camera is fixed in the scene, because the robot usually takes a long time to complete a grasp operation, the accuracy of grasp detection is more important than real-time. 
But if the camera is installed above the wrist, real-time is more important. 
We assume that the camera is fixed in the scene and always facing the table where the target is located, and conduct research to improve the accuracy and stability of the grasp detection.

Most of the current research focuses on the grasping of parallel plate gripper, Jiang et al.\cite{1} used a five-dimensional grasp representation as the grasp model to be detected for the first time, which is a simplification of the seven-dimensional grasp representation in the real world, and many people have achieved good results based on it.
However, the five-dimensional grasp representation is only suitable for parallel plate gripper that can grasp objects symmetrically, not suitable for unsymmetrical three-finger gripper.
The position of the two parallel plates can be interchanged when the parallel plate gripper grasps the object, but the unsymmetrical three-finger gripper cannot be interchanged on the single-finger side and the two-finger side when grasping because the space for single-finger side may be too small for two-finger side.
In addition, the five-dimensional rectangle represents grasping is a macro model, which is easier to understand, but it can not represent grasping authentically. 
We observe human grasping behavior and find that humans always grasp a certain region on the target to grasp the target.
Similar to human behavior, we represent feasible grasping on the target as certain regions, each of which has the same grasp angle and width. 
For the above two reasons, we designed a pixel-level grasp representation, oriented base-fixed triangle grasp representation.
It allows us to detect grasping using only RGB images and can be mapped to the seven-dimensional grasp representation in the real world. 

We introduce a novel one-stage network named SGDN for detecting good robotic grasping for unsymmetrical three-finger gripper using oriented base-fixed triangle grasp representation.
Based on the idea of semantic segmentation, DeepLabv3+ network \cite{13} is used as the main architecture of SGDN which is the best architecture in the current semantic segmentation community, we construct SGDN to predict each dimension of the oriented base-fixed triangle simultaneously on RGB images.
We relabeled the Cornell Grasp Dataset at the pixel level based on the new grasp representation.
Experiments show, our model achieves an accuracy of 96.8\% and 92.27\% on image-wise split and object-wise split respectively, and obtains accurate predictions consistent with the state-of-the-art methods.
Our real-time, generative grasping pipeline is shown in Fig.~\ref{fig1}.

Our work mainly has two contributions:
\begin{itemize}
	\item A oriented base-fixed triangle representation is proposed, which is suitable for unsymmetrical three-finger gripper to perform grasping. 
	It is also a pixel-level grasp representation method which better characterizes the realistic grasp.
	\item A novel grasp detection network named SGDN is proposed based on the idea of semantic segmentation, and SGDN obtains accurate predictions consistent with the state-of-the-art methods.
\end{itemize}

This paper is organized as follows. 
Section \uppercase\expandafter{\romannumeral2} discusses related grasp detection methods.
Section \uppercase\expandafter{\romannumeral3} presents the oriented base-fixed triangle representation and provides a detailed description of relabeling of Cornell Dataset. 
Section \uppercase\expandafter{\romannumeral4} presents the structure of SGDN.
Section \uppercase\expandafter{\romannumeral5} demonstrates detailed experiments setup.
We present our results in Section \uppercase\expandafter{\romannumeral6}, then conclude in \uppercase\expandafter{\romannumeral7}.

% 相关工作
\section{Relate work}
The goal of grasp detection is to find a proper posture through the visual information of the target to be grasped, so that the gripper can stably grasp the target when closing the gripper in this posture.
The proposed methods can be roughly divided into two categories: analytic methods and empirical methods \cite{17,18}.
Analytic methods use mathematical and physical models of geometry, kinematics and dynamics to calculate grasping that are stable \cite{19}, but tend to not transfer well to the real world due
to the difficultly in modeling physical interactions between a manipulator and an object \cite{20}.
In contrast, empirical methods does not need to know the three-dimensional model of the object to be grasped. It constructs the grasp model from the known object and uses this model to detect the grasp posture of the unknown object \cite{21, 22, 23}.
Now, method based on deep learning, detects grasp configuration on the plane firstly and then maps the planar grasp configuration to the grasp posture in the world coordinate system, is followed by many researchers and performs state-of-the-art grasp detection results over conventional methods.

The early research of deep learning mainly focused on target classification and target detection \cite{24, 25}. 
With these tasks as carriers, many excellent image feature extraction architectures were proposed \cite{16, 26, 27}.
Due to the many similarities between grasp detection and target detection, many researchers have tried to modify the network originally designed for target detection to detect grasping, and achieved inspiring results \cite{7, 28, 6}.
This has promoted the extensive research of deep learning in the robotic grasp detection community.
Using deep learning to predict grasping has several benefits.
First, it provides a certain generalization ability for grasp detection, thus making it applicable to previously unseen objects. Second, it is trained on data, making it possible to add more features in order to increase the performance. 

The goal of grasp detection is to predict a planar grasp representation used to characterize the grasp configuration, which generally includes grasp position, grasp angle, and grasp width.
Saxena et al. \cite{3} used supervised learning to predict a suitable grasp point from the image and successfully extended it to new targets.
% 图2 三角形表示+非对称三指灵巧手
\begin{figure}[htbp]
	\centerline
	{\includegraphics[scale=0.45]{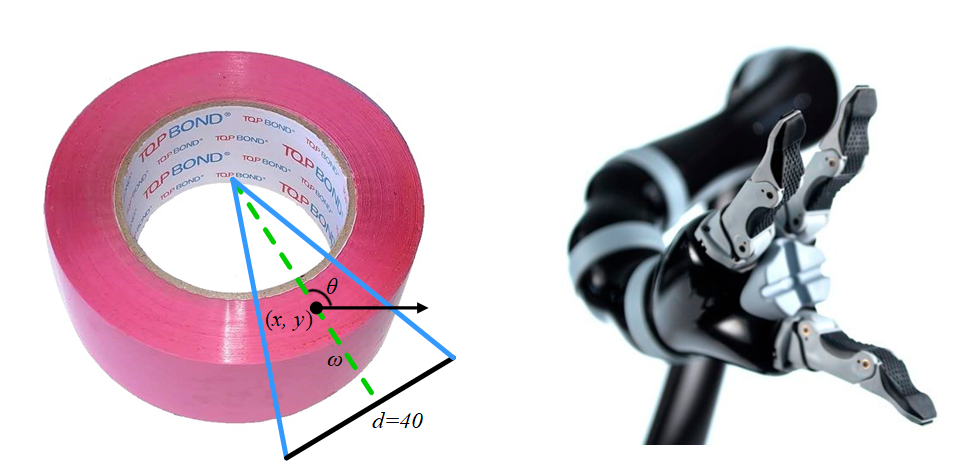}}
	\caption{\textbf{left:} A 4D grasp representation based on the oriented base-fixed triangle. A grasp is defined by triangle center coordinates $(x, y)$, angle $\theta$ of height relative to the horizontal axis, height $\omega$ of triangle and base $d$ of triangle. \textbf{right:} The unsymmetrical three-finger gripper we use.}
	\label{fig2}
\end{figure}
However, it is not possible to perform grasping without the grasp angle and width.
Le et al. \cite{4} went further than Saxena, suggesting that a pair of points is used to represent the grasping. In addition to the location of the grasping, it also includes the grasp angle and the width of the gripper opening.
Jiang \cite{1} reduced the dimensionality of the seven-dimensional grasp posture (the 3D location, the 3D orientation and the distance between the two fingers) in the real environment to obtain a simplified five-dimensional rectangular grasp representation, in fact excluding the approach vector of the gripper.
A lot of work has been done on the basis of five-dimensional rectangular grasp representation \cite{29, 30} and set the approach vector to be perpendicular to the table \cite{11, 8} or the surface normal of the grasp point \cite{23}.
Xu \cite{2} reduced the dimensionality of the five-dimensional rectangular grasp representation again, and proposed oriented diameter circle representation, which indicates the actual grasping more faithfully. However, neither oriented diameter circle nor the five-dimensional rectangle is suitable for an unsymmetrical three-finger gripper.

The research route of neural network in grasp detection is similar to target detection, from single rectangular regression to anchor regression \cite{5, 6}, from single-modality to multi-modality \cite{7, 8, 9}, from single-scale to multi-scale \cite{10} , from image level to pixel level \cite{10, 11}.
Xu et al. \cite{2} divided the input RGB images into $13\times13$ grids, and each grid regressed the oriented diameter circle whose center point falls within the grid through the GraspCNN network. 
GraspCNN was combined with the YOLO network to complete grasp detection and target detection simultaneously.
Zhou et al. \cite{5} used ResNet-50 as the feature extraction network and used anchor mechanism \cite{12} to predict the five-dimensional rectangle grasp representation, which greatly improved the prediction accuracy and finally obtained accuracy of 97.74\% in Cornell Grasp Dataset.
Asif et al. \cite{10} overcome limitations of the individual models by combined CNN structures with layer-wise feature fusion and produces grasping and their confidence scores at different levels of the image hierarchy (i.e., global-, region-, and pixel-levels).
Morrison et al. \cite{11} regarded grasp detection as a semantic segmentation problem. The proposed GGCNN predicts the grasp probability, grasp angle and width of each pixel from the depth image through a fully convolutional neural network (FCN). GGCNN is very real-time because it has very few network parameters. And GGCNN is the closest network we can find to our network.

Most researchers adopt the rectangle-based method, with two edges corresponding to the gripper plates.
However, the rectangle representation is not suitable for an unsymmetrical three-finger gripper.
Moreover, most researchers directly use the neural network that was originally designed for target detection for grasp detection. 
In fact, the two fields are very different. In target detection, the label of each target is only a rectangular boundingbox, but feasible grasp can not be fully characterized by some rectangular boxes.
To address the above problems, this paper uses a oriented base-fixed triangle representation that is suitable for an unsymmetrical three-finger gripper to characterize the planar grasp configuration, and uses a novel network to simultaneously predict the grasp position, angle and width at pixel-levels based on the idea of semantic segmentation.

% 问题描述
\section{Problem description}
\subsection{Grasp Representation}
The oriented base-fixed triangle representation is inspired from human grasp behavior.
Before grasp an object, humans will first consider the position on the object that can be grasped and the posture of the hand when grasping.
In the image plane, we represent the position on the target that can be grasped as the coordinates of the pixel, and map the posture of the hand during grasping to the grasp angle on the plane and the width of the gripper opening (i.e., grasp width).
Therefore, a planar grasp representation that can authentically characterize the planar grasp should take the pixel as the core, and include both grasp angle and width.
In addition, the unsymmetrical three-finger gripper we use is very different from the parallel plate gripper.
The parallel plate grabber can exchange the position of two parallel plates during grasping, which is expressed in mathematics: grasp angle $\theta \in [0, \pi)$.
In contrast, the three fingers of an unsymmetrical three-finger gripper are divided into single-finger side and two-finger side. The two-finger side requires more space than the single-finger side when grasping, and the two sides cannot be interchanged, that is, $\theta \in [0, 2\pi)$.
In order to solve the above problem, we designed an oriented base-fixed triangle representation, which is represented as:
\begin{equation}
G = \{x, y, \omega, \theta, d \}
\label{eq1}
\end{equation}
with $(x, y)$ denoting center of the height of triangle, $\omega$ denoting the height of triangle and $\theta$ denoting the angle of the height relative to horizontal axis of image, $d$ denoting the base of triangle. 
The apex and base of the triangle represent the one- and two-finger side of the unsymmetrical three-finger gripper respectively.
We assume that the relative position between the two fingers on the two-finger side is constant when grasping, that is, $d$ is a fixed value ($d=40$), the unit is pixel.
Fig.~\ref{fig2} shows an example of this grasp representation and the unsymmetrical three-finger gripper that we use.
The oriented base-fixed triangle is a simplification of the full 7D gripper configuration (the 3D position, 3D orientation and the gripper opening width) and can be projected back to the full 7D gripper configuration.
The center $(x, y)$ can be used to obtain the 3D grasp position from point cloud; the angle $\theta$ and surface normal of grasp position are used to obtain the 3D orientation; the gripper opening width can be calculated using the intrinsic parameters of camera and height $\omega$ of triangle.

Compared with some previous representations for robotic grasping, the oriented base-fixed triangle representation has two advantages:
% 图3 标注示例
\begin{figure}[htbp]
	\centerline
	{\includegraphics[scale=0.345]{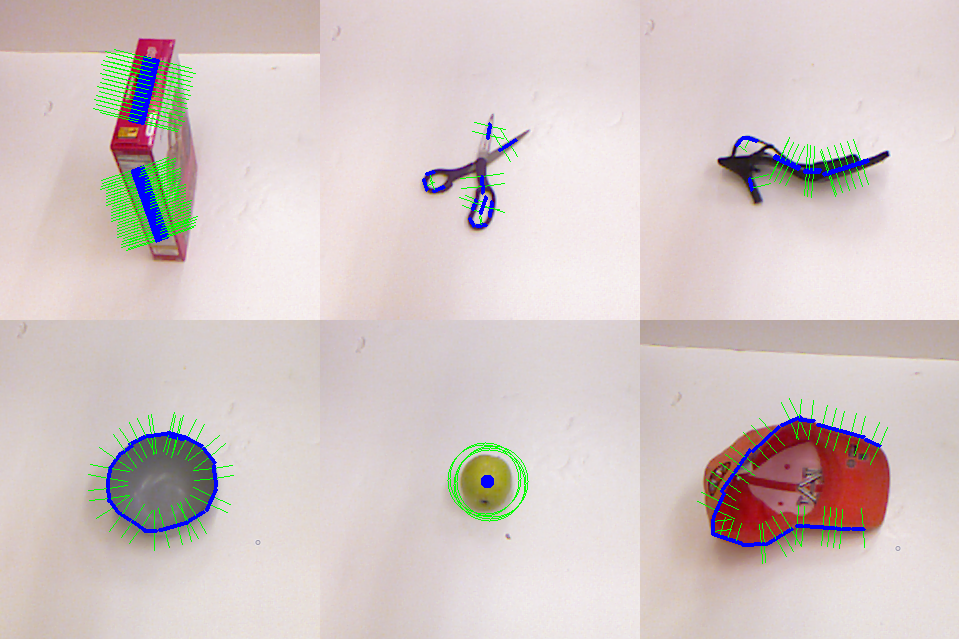}}
	\caption{Some examples of dataset labeling. The points in the blue region are all grasp points.
		Draw a green line with each grasp point as the end point. The angle between the green line and the horizontal axis is grasp angle. The length of the green line is half of the grasp width. Note that only one green line is drawn for the grasp points that cannot be symmetrically grasped, and two green lines in opposite directions are drawn for the grasp points that can be symmetrically grasped. A green circle indicates that there is no restriction on the grasp angle.}
	\label{fig3}
\end{figure}
% 图4 三角形 矩形 IOU对比
\begin{figure}[htbp]
	\centerline{\includegraphics[scale=1.0]{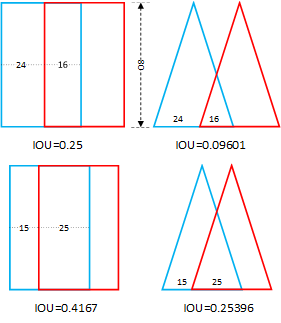}}
	\caption{The IOU comparison between triangle representation and rectangle representation. When the predicted grasp parameters are the same, the calculated IOUs based on different grasp representations vary greatly. The ground truth is shown in blue and prediction results are shown in red.}
	\label{fig4}
\end{figure}
\begin{itemize}
	\item Grasping triangle is more suitable for unsymmetrical three-finger grippers. it make grasping more stable when grippers grasp objects with complex shapes.
	\item Grabbing triangle is a pixel-level representation method, which can provide more accurate labeling results and improve the prediction accuracy.
	The details of the labeling will be introduced in section \uppercase\expandafter{\romannumeral3}-B.
\end{itemize}

\subsection{Relabel Cornell Grasp Dataset}
To train our network, we create a dataset from the Cornell Grasp Dataset \cite{1}. The Dataset contains 885 RGB-D images of real objects, with 5110 human-labelled positive and 2909 negative grasps. 
We only keep the RGB images in the dataset for annotation, and delete all other data.
Usually, when people grasp targets, they do not grasp a certain point precisely, but grasp a certain region.
Similarly, we mark regions that can be grasped on the target according to experience. All points in a region are grasp points, and these grasp points have the same grasp angle and width.
The position of each grasp point corresponds to $(x, y)$ in the oriented base-fixed triangle representation, and the grasp angle and width correspond to $\theta$ and $\omega$ respectively, that is, each grasp point corresponds to an oriented base-fixed triangle representation.
When the space around the target's graspable region is large, it is also allowed to swap the positions of the single-finger side and the two-finger side of the unsymmetrical three-finger gripper, that is, one grasp point may have two grasp angles.
For round objects, the grasp angle is not even constrained.
Put another way, each grasp point may correspond to multiple oriented base-fixed triangle representation.
In addition, we add grasp confidence $q$ to the oriented base-fixed triangle representation, $q$ represents the probability that this oriented base-fixed triangle can be used as a grasp representation to perform grasp.
Fig.~\ref{fig3} shows some examples of dataset labeling.

In order to facilitate the training of the network, we quantify the labeled data.

\textbf{Grasp Confidence:} We treat grasp confidence of each point as a binary label and set the points in already marked regions a value of 1, all other points are 0.

\textbf{Angle:} We compute the angle of each grasp triangle in the range $[0, 2 \pi)$ and discretize it from 0 to $k$, $k \in [36, 72, 120]$.
We will compare the effects of different $k$ in section \uppercase\expandafter{\romannumeral6}.

\textbf{Width:} Since the maximum value of all the grasp widths of the label is close to 150, 
we scale the values of $\omega$ by $\frac{1}{150}$ to put it in the range $[0, 1)$. The physical gripper width can be calculated using the parameters of the camera and the measured depth.

The grasp angle and width of all points except the labeled grasp points are all set to 0.

\subsection{Grasp Evaluation Metric}
To evaluate the predicted grasp, the most commonly used method now is that the predicted grasp is correct if both:
\begin{itemize}
	\item The difference between the predicted grasp angle $\theta$ and the ground-truth grasp angle is less than $30^{\circ}$.
	\item The Intersection Over Union (IOU) of the predicted rectangle and ground-truth grasp rectangle is greater than 0.25.
\end{itemize}

Since the grasp representation we used is oriented base-fixed triangle instead of rectangle, we compare the IOU of triangle representation with the IOU of rectangle representation under different prediction results. 
Fig.~\ref{fig4} shows the comparison results.
The results show that a small prediction error can make the evaluation based on the triangle representation very poor.
Predictive grasp that meets the evaluation metric based on rectangle representation is far from satisfying the evaluation metric based on triangle representation.
In order to minimize the unfairness when compared with the previous algorithm, we finally use the evaluation metric based on rectangle representation.

\section{Network}
% 图5 
\begin{figure*}[htbp]
	\centering
	{\includegraphics[scale=0.45]{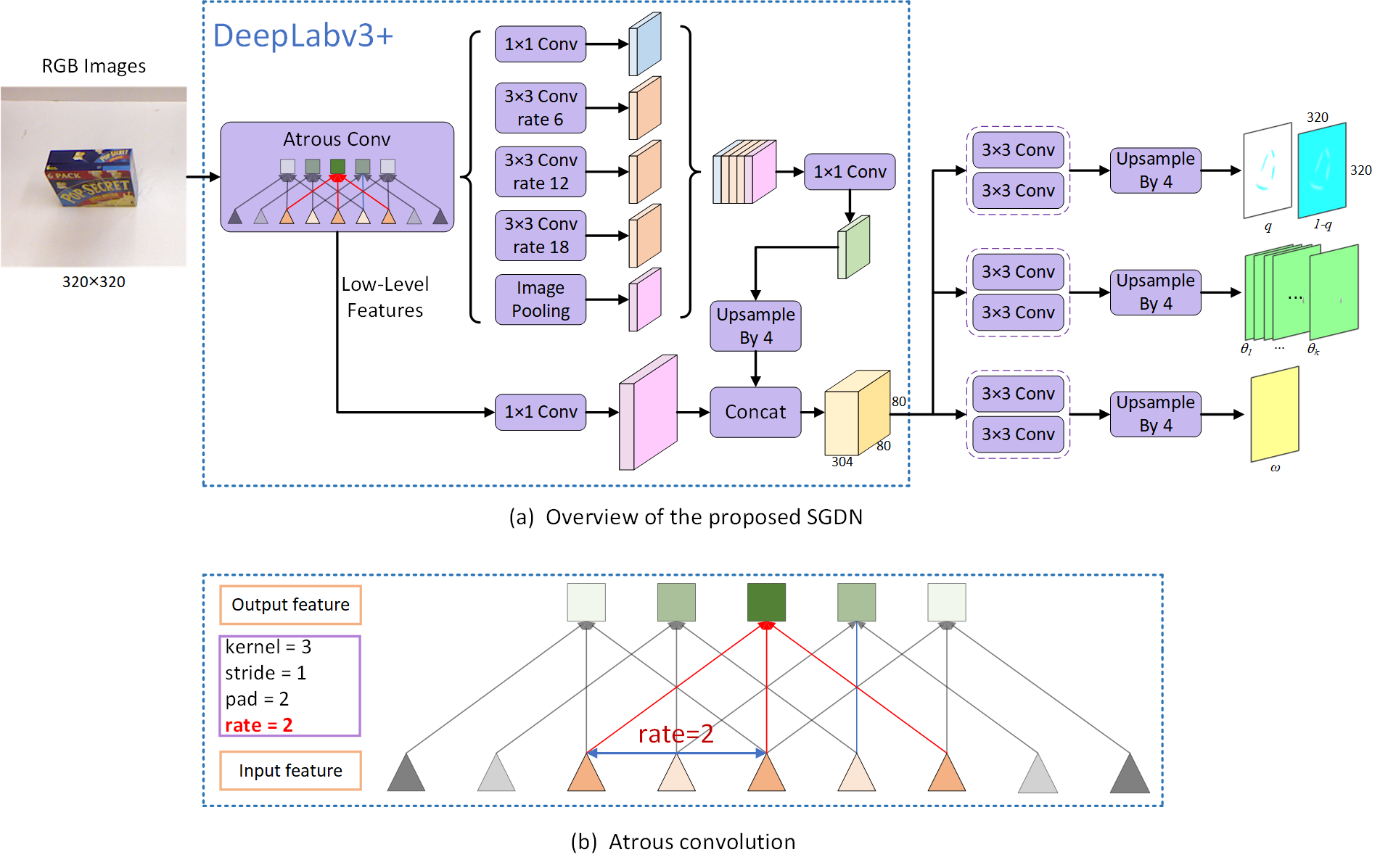}}
	\caption{(a) The feature extractor with DeepLadv3+ as the backbone extracts features from the RGB image, followed by grasp predictor predicts feasible grasp triangle. (b) Dense feature extraction with atrous convolution with rate of 2, applied on a high resolution input feature map.}
	\label{fig5}
\end{figure*}
The prediction goal of the network we are going to build is to input an RGB image and predict the potential grasp points and the grasp angle and width at each grasp point.
All potential grasp points in an image will converge into multiple scattered regions, and the grasp points in each region have the same grasp angle and width.
In other words, if we don't consider the grasp angle and width for the time being, the prediction goal of the network is to input an RGB image and predict the potential grasp region in the image.
Further, the potential grasp regions and other regions in the image are divided into two categories according to the grasp confidence $q$, and the network is used to classify each pixel in the image as belonging to the grasp region or other regions.
This is actually semantic segmentation, except that there are only two categories.
If we consider the grasp angle and width again, the prediction goal of the network is to predict the grasp angle and width of each pixel while semantic segmentation.

Starting from the idea of semantic segmentation, we consider improving the existing semantic segmentation network to meet our needs instead of rebuilding the network.
The DeepLabv3+ network proposed by Chen et al. \cite{13} achieved the best results on the PASCAL VOC 2012 semantic segmentation competition.
We use the DeepLabv3+ as the main architecture of our network and have modified it to meet our needs.

\subsection{Model Architecture}
One of the highlights of DeepLabv3+ is the use of atrous convolution \cite{14}.
Atrous convolution is obtained by upsampling the original convolution filter and introducing zeros in between filter values, it is a convolution filter `with holes'. 
Atrous convolution allows us to explicitly control the resolution at which feature responses are computed within Deep Convolutional Neural Networks. It also allows us to effectively enlarge the field of view of filters to incorporate larger context without increasing the number of parameters or the amount of computation. 
A sample of atrous convolution with rate of $2$ is shown in Fig.~\ref{fig5}.

DeepLabv3+ combined the advantages from spatial pyramid pooling module and encode-decoder structure.
The encoder module encodes multi-scale contextual information by applying atrous convolution at multiple scales, while the simple yet effective decoder module refines the segmentation results along object boundaries.
ImageNet-1k \cite{15} pretrained ResNet-101 \cite{16} is used as DeepLabv3+ backbone.
For more detail about DeepLabv3+, please refer to \cite{13}.

We modified the original DeepLabv3+ architecture for our grasp detection.
The last convolutional layer and upsampling layer of DeepLabv3+ are deleted, and the other parts are used as a larger feature extractor.
Taking $320\times320$ pixels RGB image as input, the feature extractor produces 304 feature maps of size $80\times80$, as illustrated in Fig.~\ref{fig5}.

The feature maps are input into a grasp predictor.
The grasp predictor has three components: grasp confidence predictor, grasp angle predictor and grasp width predictor. 
Each of them includes two convolutional layers of size $3\times3$ and an upsampling layer by a factor of 4, the size of the final feature maps are all $320\times320$, as same as the input images.
At each of the $320\times320$ locations of feature maps, the grasp confidence predictor produces 2 output values, the probability of being graspable and the probability of not being graspable. 
The grasp angle predictor produces $k$ output values, and the $i_{th}$ value represents the probability that the grasp angle belongs to class $i$, $i=0, ..., k-1$.
The grasp width predictor produces 1 output value, that is, grasp width.
Our model is one-stage detector, which directly predict grasps from the feature maps produced by feature extractor.

From another perspective, our grasp confidence predictor predicts oriented base-fixed triangle centers by determining whether the point location is graspable. At the same time, grasp angle and width predictor refine  oriented base-fixed triangle scale and orientation.

\subsection{Loss Function}
Prediction grasp confidence is a binary classification problem, so we use the softmax cross-entropy function as the loss function of predicting grasp confidence.
Since the unsymmetrical three-finger gripper may have more than one selectable grasp angle when grasping objects, predicting the grasp angle is a multi-label multi-classification problem, so we use the sigmoid cross-entropy function as the loss function for predicting grasp angle.
Predicting grasp width is a regression problem, so and the mean square error function is selected as the loss function.

\section{Experiment setup}
\subsection{Dataset}
In order to compare with other algorithms, our models are trained and tested on Cornell Grasp Dataset. 
We relabeled the dataset, and the details of the labeling are introduced in section \uppercase\expandafter{\romannumeral3}-B.

Like previous works, we randomly selected $75\%$ as the training set in the dataset, and the remaining $25\%$ as the test set.
The test methods are divided into the following two:
\begin{itemize}
	\item \textbf{Image-wise split} divides the images into training set and test set at random. This aims to test the generalization ability of the network to new position and orientation of an object it has seen before.  
	\item \textbf{Object-wise split} divides the dataset at object instance level. All the images of an instance are put into the same set. This aims to test the generalization ability of the network to 
	new object. 
\end{itemize}

% 数据增强待补充
A deep neural network has complicated structure and many parameters, which requires a large number of manually labeled dataset.
However, collecting a large number of dataset is very tedious work.
In order to improve training efficiency and make full use of existing data sets, researchers generally adopt two methods.
First, a pre-trained image classification network is used as a feature extractor, and then fine-tune the network parameters on a dedicated dataset.
Second, data augmentation techniques is used to expand existing dataset.

Compared with other datasets in deep learning, the Cornell Grasp Dataset is a small dataset. 
In addition to using the ResNet-101 network pre-trained on ImageNet as a feature extractor, we also performed data enhancement on the Cornell Dataset in multiple ways.
We take the average of all the marked grasp points position as the center, and cut the input image to $320\times320$ pixels.
The cropped image block rotates clockwise, and the rotation angle is randomly generated in the range $[0, 2\pi)$.
Then, the image blocks are zoomed randomly with the midpoint of the image block as the center point.
Finally, we put the image into the network at the resolution of $320\times320$.
Our augmentation is implemented online, which means every input image is a new image from pixel-level. 

\subsection{The Best Grasp Representation}
In order to find the optimal grasp triangle, we set a grasp confidence threshold, traverse each coordinate in the feature map output by the grasp confidence predictor, and record the points whose grasp confidence is greater than the threshold as candidate grasp points.
Among the candidate grasp points, the grasp point with the highest grasp confidence is taken as the optimal grasp point.
According to the grasp angle and width prediction value corresponding to the coordinates of the optimal grasp point, the optimal oriented base-fixed triangle is calculated and used to implement the grasp operation.
If the number of candidate grasp points is zero, adjust the camera's shooting angle and re-detect.

\subsection{Implementation Details }
SGDN is implemented with Torch for its flexibility. 
For training and testing, our models run on a single NVIDIA TITAN-X. 
For each of the models we tested, we employ the same training regimen. 
There are a total of about 60 million parameters in the network, the batch size is set as 12 due to the memory limitation of the graphics card, and the memory required for a forward propagation is about 8G. 
Each model is trained end-to-end for 1500 epochs. 
We use adam optimizer to optimize SGDN, and lr decays stepwise at a rate of 0.5 times in the range $[200, 500, 800, 1000]$ of epochs.

\section{Results}
% 表二
\begin{table*}[htbp]
	\caption{Performance of different algorithms on Cornell Grasp Dataset}
	\begin{center}
		\begin{tabular}{|c|c|c|c|c|c|}
			\hline
			\multirow{2}*{\textbf{Approach}} & \multirow{2}*{\textbf{Year}} & \multirow{2}*{\textbf{Algorithm}} & \multicolumn{2}{c|}{\textbf{Accuracy (\%)} } & \multirow{2}*{\textbf{speed (fps)}} \\
			\cline{4-5}
			& & & \textbf{image-wise} & \textbf{object-wise} & \\ 
			\hline
			\makecell[l]{Jiang et al. \cite{1}} & 2011 & \makecell[l]{Fast Search} & 60.5 & 58.3 & 0.02 \\
			\hline
			\makecell[l]{Lenz et al. \cite{8}} & 2015 & \makecell[l]{SAE, struct. reg. Two stage} & 73.9 & 75.6 & - \\ 
			\hline
			\makecell[l]{Kumra et al. \cite{7}} & 2017 & \makecell[l]{ResNet-50× 2, Multi-model Grasp Predictor} & 89.21 & 88.96 & 16.03 \\
			\hline
			\makecell[l]{Zhang et al. \cite{23}} & 2019 & \makecell[l]{ROI-GD, ResNet-101, RGB} & 93.6 & 93.5 & 25.16 \\
			\hline
			\makecell[l]{\multirow{2}*{Chu et al. \cite{6}}} & \multirow{2}*{2018} & \makecell[l]{VGG-16, Deep Grasp, 3 scales and 3 aspect ratios} & 95.5 & 91.7 & 17.24 \\
			\cline{3-6}
			& & \makecell[l]{ResNet-50, Deep Grasp, 3 scales and 3 aspect ratios} & 96.0 & \textbf{96.1} & 8.33 \\
			\hline
			\makecell[l]{Xu et al. \cite{2}} & 2019 & \makecell[l]{GraspCNN, oriented diameter rectangle} & 96.5 & - & 50 \\
			\hline
			
			\makecell[l]{\multirow{2}*{\textbf{Ours}}} & \multirow{2}*{2020} & \makecell[l]{Oriented base-fixed triangle, 72 grasp angle classes} & 96.35 & 95.45 & \textbf{52.34} \\
			\cline{3-6}
			& & \makecell[l]{Oriented base-fixed triangle, 120 grasp angle classes} & \textbf{96.8} & 92.27 & 51.6 \\
			\hline
		\end{tabular}
		\label{tab2}
	\end{center}
\end{table*}
\begin{table}[htbp]
	\caption{Accuracy under different grasp angle classification}
	\begin{center}
		\begin{tabular}{|c|c|c|c|}
			\hline
			\makecell[c]{Grasp Angle \\ Classification} & \textbf{Image-wise} & \textbf{Object-wise} & \textbf{speed (fps)}\\ 
			\hline
			36 & 95.89\% & 91.82\% & \textbf{52.99} \\ 
			\hline 
			72 & 96.35\% & \textbf{95.45\%} & 52.34 \\ 
			\hline
			120 & \textbf{96.8\%} & 92.27\% & 51.6 \\ 
			\hline
		\end{tabular}
		\label{tab1}
	\end{center}
\end{table}
In grasp angle predictor of SGDN, grasp angle prediction is regarded as a multi-label multi-classification problem. We tested the impact of using different grasp angle classification numbers on the prediction results.
The details of testing the different grasp angle classification numbers are shown in Table~\ref{tab1}.
In the image-wise mode, the prediction accuracy increases with the number of grasp angle classifications, and the maximum accuracy is 96.8\%. In the object-wise mode, when the number of grasp angle classifications is 72, the prediction accuracy is the highest, which is 95.45\%. A larger number of grasp angle classifications will reduce the error of each category and make the prediction result closer to the real situation. At the same time, the greater the number of grasp angle classifications, the more network parameters and lower fps.

% 图6 预测示例
\begin{figure}[htbp]
	\centerline
	{\includegraphics[scale=0.43]{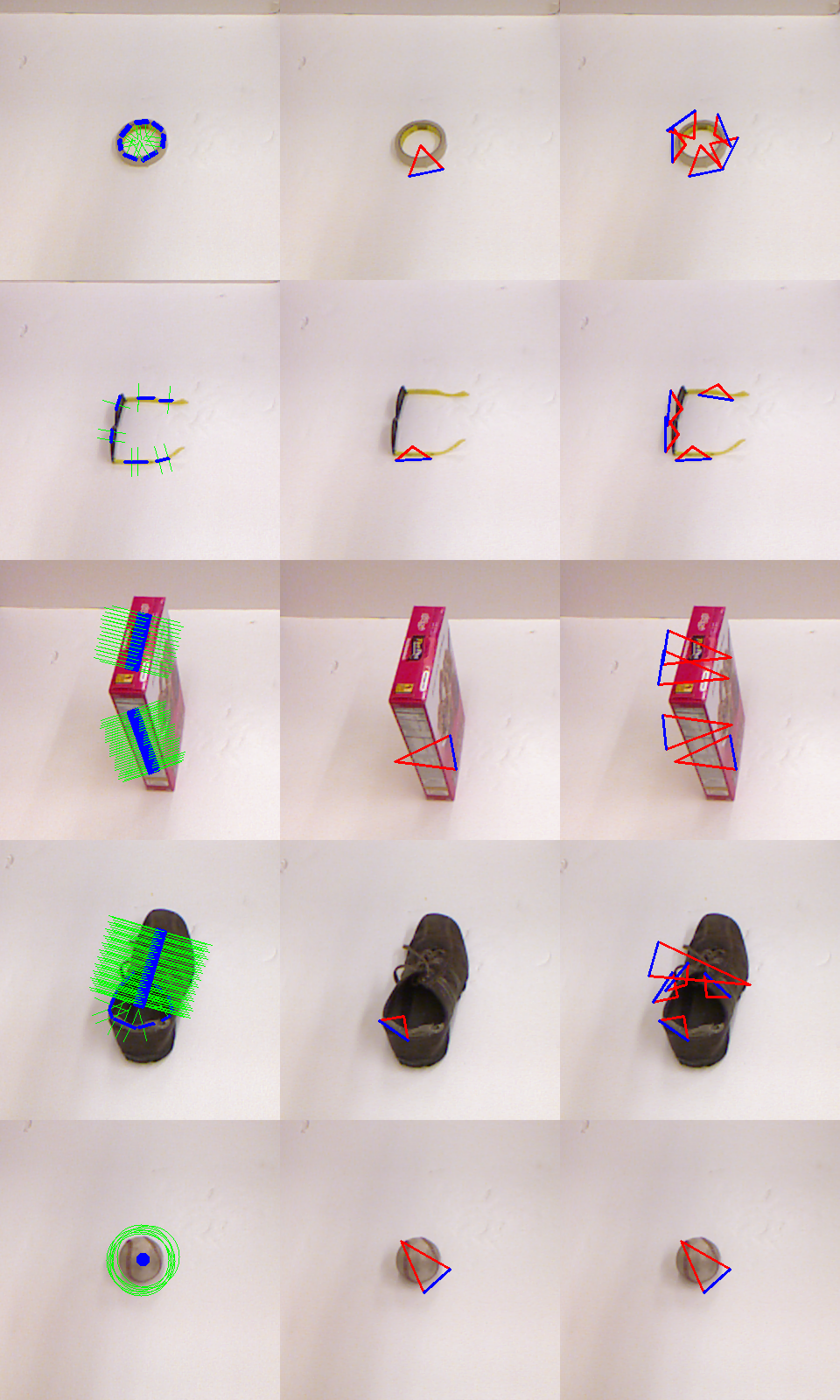}}
	\caption{Detection results in Cornell Grasp Dataset. The first column is ground-truth oriented  rectangles. The second column is the visualization of Top1 grasp detection result. The third column is the results of multi grasps.}
	\label{fig6}
\end{figure}
% 图7 预测错误示例
\begin{figure*}[htbp]
	\centerline
	{\includegraphics[scale=0.3]{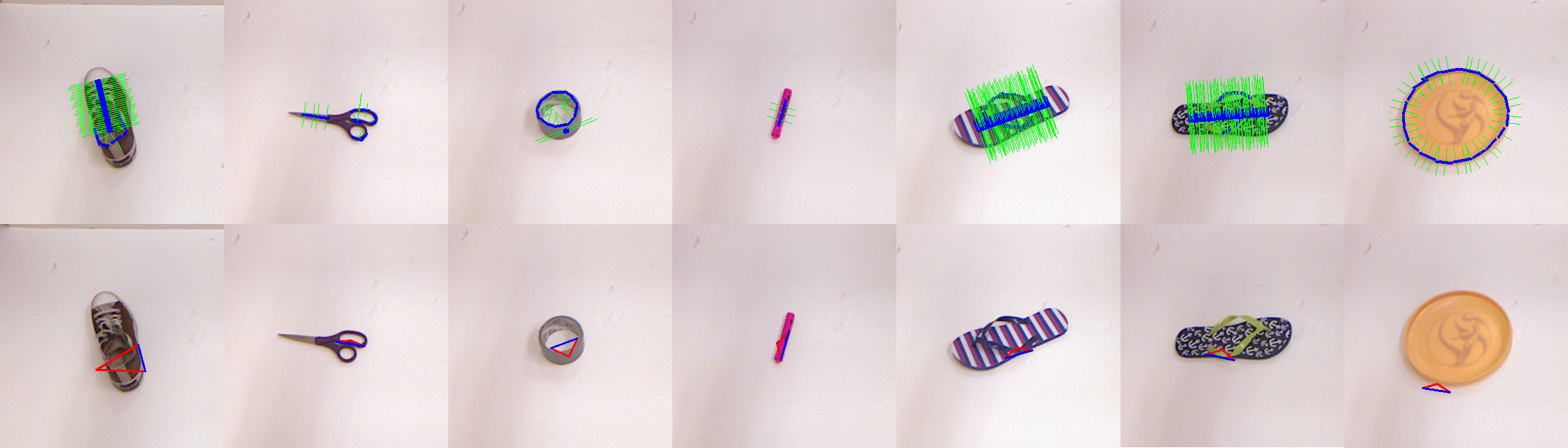}}
	\caption{Unsuccessful detection results. The first row is the labels. The second row is the detection results.}
	\label{fig7}
\end{figure*}
% 图8 实际场景测试\textsl{}
\begin{figure}[htbp]
	\centerline
	{\includegraphics[scale=0.43]{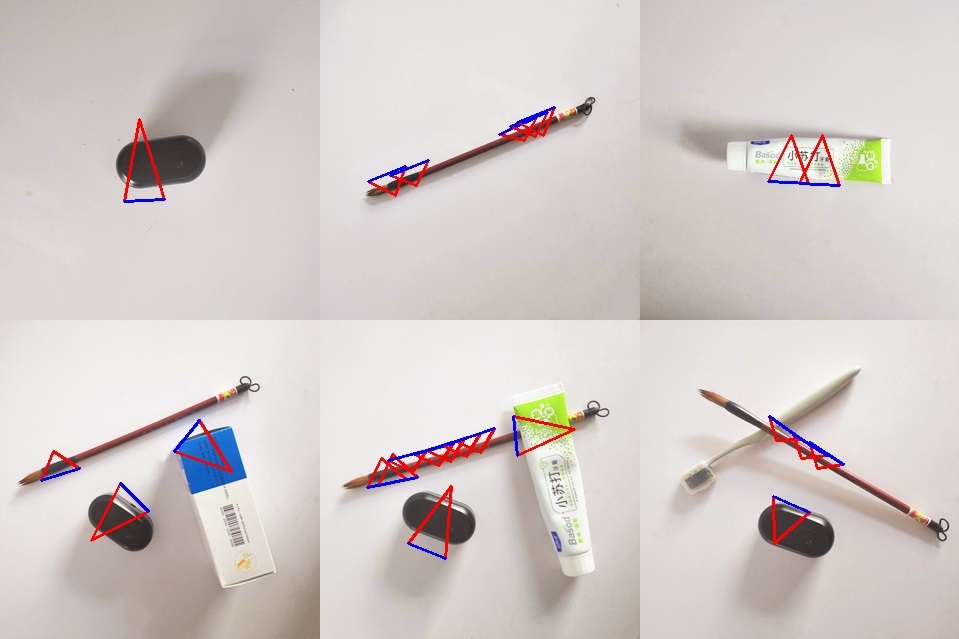}}
	\caption{Visualization of detection in more realistic and complex scene}
	\label{fig8}
\end{figure}
The closest we found to our work is GGCNN \cite{11}, but since GGCNN only shows the grasp success rate in the actual running scenario, but does not show the prediction accuracy rate on the Cornell Dataset, it is impossible to compare.
However, our method is compared with some previous methods, the comparison options include prediction accuracy and fps in image-wise and object-wise modes. The details of the comparison are shown in Table~\ref{tab2}.
Since the oriented base-fixed triangle representation we used is newly proposed, the comparison with previous work is not rigorous enough, but roughly, we use the method based on semantic segmentation to obtain accurate predictions consistent with the state-of-the-art methods.
It can be seen from Table~\ref{tab2} that most of the previous networks were modified from the network designed originally for target detection, not specifically designed for grasp detection.
Although many methods have achieved good results on the Cornell Dataset, they can still be improved.
Compared with most methods that use five-dimensional rectangle grasp representation, we have achieved better results.
We think there are two reasons.
First, the oriented base-fixed triangle representation we proposed is a pixel-level representation method. 
It characterizes the features of the grasp in the actual scene more finely than the five-dimensional rectangle representation.
Second, we use the idea of semantic segmentation to detect grasp, it directly predicts the region on the object that can be grasped, rather than an abstract grabbing representation, which makes the prediction result of the network directly correspond to the actual grasp.
In contrast, most of the previous work used regression methods to predict the five-dimensional rectangle representation. 
The prediction results of the network need to be converted through various transformations to obtain the configuration that can be used for grasping, the abstraction will then lead to poor prediction.

Our method can still be improved.
First, although the main network architecture we adopt is DeepLabv3+ that has the best effect in the field of semantic segmentation, the original function of DeepLabv3+ is to detect the target region.
For grasp detection, the graspable region is often not the target region, but some thinner and smaller internal regions, which makes DeepLabv3+ unable to exert its strongest performance.
Second, the grasp angle prediction of the oriented base-fixed triangle is a multi-label multi-classification problem.
Compared with the single-label multi-classification problem, SGDN is more difficult to fully train.
In addition, our method has many parts that need to be improved. 
These parts will make the network performance better, and it is also the focus of our follow-up work.

In Fig.~\ref{fig6}, we visualize labels and detection oriented base-fixed triangles of some objects in the test set of Cornell Grasp Dataset under image-wise splitting. The first column shows the labels of the objects. The second column visualizes the Top1 detection results of these object. Multi-grasp detection results are demonstrated in the third column. 
Since the network prediction results are pixel-level, in order to facilitate viewing, we have drawn multi-grasp detection results based on the local peaks of the grasp confidence results.
All the grasp triangles in the third column have a graspable score over 0.5. 
In the multi-grasp detection results, the predicted grasp covers almost all the labeled grasp regions, which shows that the idea based on semantic segmentation has a very high feasibility.

Under the image-wise split accuracy of 96.8\%, we has seven unsuccessful detections in total. All these false detections results are shown in Fig.~\ref{fig7}. The first row is the labels and the second row is the detection results. 
In the first four results, the coordinates of the grasp points are predicted correctly, but the grasp angle and width are incorrect; the prediction results of the fifth and sixth ones are still feasible although they do not satisfy the metric; the last grasp point is predicted incorrectly, but the angle and width predictions are correct.
This shows that our label is not complete. 
How to optimize the three predictors at the same time is also the focus of the next improvement.

To test our model on more realistic and complex scenes, we test SGDN, which is trained on image-wise split, with some pictures where the objects overlap with each other. The results is shown in Fig.~\ref{fig8}. Some categories (writing brush, headphone box) never appear in Cornell Grasp Dataset. Despite the occlusion, our model still has good performance under more realistic and complex scene. Besides, our model successfully predicts grasp for unseen objects. 

\section{Conclusion}
We represent SGDN, a new architecture for grasp detection based on the idea of semantic segmentation using RGB images, and we designed oriented base-fixed triangle grasp representation for asymmetric three-finger gripper.
On Cornell Grasp Dataset, SGDN outperforms current state-of-the-art model. Furthermore, SGDN can predict feasible grasping in a multi-object stacking scene.

Our future work will focus on improving operational efficiency and accuracy of SGDN, while collecting a larger dataset for tools commonly used in factories. In addition, we will also focus on task-oriented grasping.

% 参考文献
\bibliographystyle{IEEEtran}
\bibliography{reference.bib}

% Generated by IEEEtran.bst, version: 1.12 (2007/01/11)
\begin{thebibliography}{10}
\providecommand{\url}[1]{#1}
\csname url@samestyle\endcsname
\providecommand{\newblock}{\relax}
\providecommand{\bibinfo}[2]{#2}
\providecommand{\BIBentrySTDinterwordspacing}{\spaceskip=0pt\relax}
\providecommand{\BIBentryALTinterwordstretchfactor}{4}
\providecommand{\BIBentryALTinterwordspacing}{\spaceskip=\fontdimen2\font plus
\BIBentryALTinterwordstretchfactor\fontdimen3\font minus
  \fontdimen4\font\relax}
\providecommand{\BIBforeignlanguage}[2]{{%
\expandafter\ifx\csname l@#1\endcsname\relax
\typeout{** WARNING: IEEEtran.bst: No hyphenation pattern has been}%
\typeout{** loaded for the language `#1'. Using the pattern for}%
\typeout{** the default language instead.}%
\else
\language=\csname l@#1\endcsname
\fi
#2}}
\providecommand{\BIBdecl}{\relax}
\BIBdecl

\bibitem{1}
{Yun Jiang}, S.~{Moseson}, and A.~{Saxena}, ``Efficient grasping from rgbd
  images: Learning using a new rectangle representation,'' in \emph{2011 IEEE
  International Conference on Robotics and Automation}, 2011, pp. 3304--3311.

\bibitem{13}
L.-C. Chen, Y.~Zhu, G.~Papandreou, F.~Schroff, and H.~Adam, ``Encoder-decoder
  with atrous separable convolution for semantic image segmentation,'' in
  \emph{Computer Vision -- ECCV 2018}, Cham, 2018, pp. 833--851.

\bibitem{17}
J.~{Bohg}, A.~{Morales}, T.~{Asfour}, and D.~{Kragic}, ``Data-driven grasp
  synthesis—a survey,'' \emph{IEEE Transactions on Robotics}, vol.~30, no.~2,
  pp. 289--309, 2014.

\bibitem{18}
A.~Sahbani, S.~El-Khoury, and P.~Bidaud, ``An overview of 3d object grasp
  synthesis algorithms,'' \emph{Robotics and Autonomous Systems}, vol.~60,
  no.~3, pp. 326 -- 336, 2012, autonomous Grasping.

\bibitem{19}
A.~{Bicchi} and V.~{Kumar}, ``Robotic grasping and contact: a review,'' in
  \emph{Proceedings 2000 ICRA. Millennium Conference. IEEE International
  Conference on Robotics and Automation. Symposia Proceedings (Cat.
  No.00CH37065)}, vol.~1, 2000, pp. 348--353 vol.1.

\bibitem{20}
C.~{Rubert}, D.~{Kappler}, A.~{Morales}, S.~{Schaal}, and J.~{Bohg}, ``On the
  relevance of grasp metrics for predicting grasp success,'' in \emph{2017
  IEEE/RSJ International Conference on Intelligent Robots and Systems (IROS)},
  2017, pp. 265--272.

\bibitem{21}
Y.~{Inagaki}, R.~{Araki}, T.~{Yamashita}, and H.~{Fujiyoshi}, ``Detecting
  layered structures of partially occluded objects for bin picking,'' in
  \emph{2019 IEEE/RSJ International Conference on Intelligent Robots and
  Systems (IROS)}, 2019, pp. 5786--5791.

\bibitem{22}
A.~{Gariépy}, J.~{Ruel}, B.~{Chaib-draa}, and P.~{Giguère}, ``Gq-stn:
  Optimizing one-shot grasp detection based on robustness classifier,'' in
  \emph{2019 IEEE/RSJ International Conference on Intelligent Robots and
  Systems (IROS)}, 2019, pp. 3996--4003.

\bibitem{23}
H.~{Zhang}, X.~{Lan}, S.~{Bai}, X.~{Zhou}, Z.~{Tian}, and N.~{Zheng},
  ``Roi-based robotic grasp detection for object overlapping scenes,'' in
  \emph{2019 IEEE/RSJ International Conference on Intelligent Robots and
  Systems (IROS)}, 2019, pp. 4768--4775.

\bibitem{24}
A.~Voulodimos, N.~Doulamis, A.~Doulamis, and E.~Protopapadakis, ``Deep learning
  for computer vision: A brief review,'' \emph{Computational Intelligence and
  Neuroscience}, vol. 2018, pp. 1--13, 02 2018.

\bibitem{25}
J.~Schmidhuber, ``Deep learning in neural networks: An overview,'' \emph{Neural
  Networks}, vol.~61, pp. 85 -- 117, 2015.

\bibitem{16}
K.~{He}, X.~{Zhang}, S.~{Ren}, and J.~{Sun}, ``Deep residual learning for image
  recognition,'' in \emph{2016 IEEE Conference on Computer Vision and Pattern
  Recognition (CVPR)}, 2016, pp. 770--778.

\bibitem{26}
K.~Simonyan and A.~Zisserman, ``Very deep convolutional networks for
  large-scale image recognition,'' \emph{arXiv 1409.1556}, 09 2014.

\bibitem{27}
C.~Szegedy, S.~Ioffe, V.~Vanhoucke, and A.~Alemi, ``Inception-v4,
  inception-resnet and the impact of residual connections on learning,'' 2016.

\bibitem{7}
S.~{Kumra} and C.~{Kanan}, ``Robotic grasp detection using deep convolutional
  neural networks,'' in \emph{2017 IEEE/RSJ International Conference on
  Intelligent Robots and Systems (IROS)}, 2017, pp. 769--776.

\bibitem{28}
H.~{Karaoguz} and P.~{Jensfelt}, ``Object detection approach for robot grasp
  detection,'' in \emph{2019 International Conference on Robotics and
  Automation (ICRA)}, 2019, pp. 4953--4959.

\bibitem{6}
F.~{Chu}, R.~{Xu}, and P.~A. {Vela}, ``Real-world multiobject, multigrasp
  detection,'' \emph{IEEE Robotics and Automation Letters}, vol.~3, no.~4, pp.
  3355--3362, 2018.

\bibitem{3}
A.~Saxena, J.~Driemeyer, J.~Kearns, and A.~Y. Ng, ``Robotic grasping of novel
  objects,'' in \emph{In Neural Information Processing Systems}, 2006, pp.
  1209--1216.

\bibitem{4}
Q.~V. {Le}, D.~{Kamm}, A.~F. {Kara}, and A.~Y. {Ng}, ``Learning to grasp
  objects with multiple contact points,'' in \emph{2010 IEEE International
  Conference on Robotics and Automation}, 2010, pp. 5062--5069.

\bibitem{29}
C.~{Yang}, X.~{Lan}, H.~{Zhang}, and N.~{Zheng}, ``Task-oriented grasping in
  object stacking scenes with crf-based semantic model,'' in \emph{2019
  IEEE/RSJ International Conference on Intelligent Robots and Systems (IROS)},
  2019, pp. 6427--6434.

\bibitem{30}
G.~{Wu}, W.~{Chen}, H.~{Cheng}, W.~{Zuo}, D.~{Zhang}, and J.~{You},
  ``Multi-object grasping detection with hierarchical feature fusion,''
  \emph{IEEE Access}, vol.~7, pp. 43\,884--43\,894, 2019.

\bibitem{11}
D.~Morrison, P.~Corke, and J.~Leitner, ``Closing the loop for robotic grasping:
  A real-time, generative grasp synthesis approach,'' 2018.

\bibitem{8}
I.~Lenz, H.~Lee, and A.~Saxena, ``Deep learning for detecting robotic grasps,''
  \emph{The International Journal of Robotics Research}, vol.~34, no. 4-5, pp.
  705--724, 2015.

\bibitem{2}
Y.~{Xu}, L.~{Wang}, A.~{Yang}, and L.~{Chen}, ``Graspcnn: Real-time grasp
  detection using a new oriented diameter circle representation,'' \emph{IEEE
  Access}, vol.~7, pp. 159\,322--159\,331, 2019.

\bibitem{5}
X.~{Zhou}, X.~{Lan}, H.~{Zhang}, Z.~{Tian}, Y.~{Zhang}, and N.~{Zheng}, ``Fully
  convolutional grasp detection network with oriented anchor box,'' in
  \emph{2018 IEEE/RSJ International Conference on Intelligent Robots and
  Systems (IROS)}, 2018, pp. 7223--7230.

\bibitem{9}
D.~{Guo}, F.~{Sun}, H.~{Liu}, T.~{Kong}, B.~{Fang}, and N.~{Xi}, ``A hybrid
  deep architecture for robotic grasp detection,'' in \emph{2017 IEEE
  International Conference on Robotics and Automation (ICRA)}, 2017, pp.
  1609--1614.

\bibitem{10}
U.~Asif, J.~Tang, and S.~Harrer, ``Densely supervised grasp detector (dsgd),''
  \emph{Proceedings of the AAAI Conference on Artificial Intelligence},
  vol.~33, pp. 8085--8093, 07 2019.

\bibitem{12}
S.~{Ren}, K.~{He}, R.~{Girshick}, and J.~{Sun}, ``Faster r-cnn: Towards
  real-time object detection with region proposal networks,'' \emph{IEEE
  Transactions on Pattern Analysis and Machine Intelligence}, vol.~39, no.~6,
  pp. 1137--1149, 2017.

\bibitem{14}
L.~{Chen}, G.~{Papandreou}, I.~{Kokkinos}, K.~{Murphy}, and A.~L. {Yuille},
  ``Deeplab: Semantic image segmentation with deep convolutional nets, atrous
  convolution, and fully connected crfs,'' \emph{IEEE Transactions on Pattern
  Analysis and Machine Intelligence}, vol.~40, no.~4, pp. 834--848, 2018.

\bibitem{15}
O.~Russakovsky, J.~Deng, H.~Su, J.~Krause, S.~Satheesh, S.~Ma, Z.~Huang,
  A.~Karpathy, A.~Khosla, M.~Bernstein, A.~Berg, and L.~Fei-Fei, ``Imagenet
  large scale visual recognition challenge,'' \emph{International Journal of
  Computer Vision}, vol. 115, 09 2014.

\end{thebibliography}

\end{document}